\documentclass{article}

\usepackage{arxiv}
\usepackage{afterpage}
\usepackage{amssymb}
\usepackage{adjustbox}
\usepackage{bibentry}  
\usepackage{setspace}
\usepackage{enumitem}
\usepackage{xcolor}
\usepackage{soul}
\usepackage{framed} 
\usepackage{multirow}
\usepackage{caption}
\usepackage{rotating}
\usepackage{colortbl}
\usepackage{subfiles}
\usepackage{amsmath}
\usepackage{hyperref}
\usepackage{arydshln}
\usepackage{tikz}
\usepackage{gensymb}
\usepackage{lscape}
\usepackage{array}
\usepackage{subfiles}
\usepackage{mdwmath}
\usepackage{mdwtab}
\usepackage{eqparbox}
\usepackage{url}
\usepackage{pdflscape}
\usepackage{caption}
\usepackage{subcaption}
\def\checkmark{\tikz\fill[scale=0.4](0,.35) -- (.25,0) -- (1,.7) -- (.25,.15) -- cycle;} 

\title{Predicting Next Useful Location With Context-Awareness: The State-Of-The-Art}

\author{
    Alireza Nezhadettehad \\
    School of Information Technology\\
    Deakin University, Burwood, Australia\\
\And
    Arkady Zaslavsky \\
    School of Information Technology\\
    Deakin University, Burwood, Australia\\
\And
    Rakib Abdur\\
    Systems Security Group\\
    Centre for Future Transport and Cities\\
    Coventry University, Coventry, UK
\And
    Siraj Ahmed Shaikh\\
    Systems Security Group\\
    Department of Computer Science\\
    Swansea University, Wales
\And
    Seng W. Loke\\
    School of Information Technology\\
    Deakin University, Burwood, Australia\\
\And
    Guang-Li Huang\\
    School of Information Technology\\
    Deakin University, Burwood, Australia\\
\And
    Alireza Hassani\\
    School of Information Technology\\
    Deakin University, Burwood, Australia\\
}

\begin{document}
\maketitle
\begin{abstract}
  Predicting the future location of mobile objects reinforces location-aware services with proactive intelligence and helps businesses and decision-makers with better planning and near real-time scheduling in different applications such as traffic congestion control, location-aware advertisements, and monitoring public health and well-being. The recent developments in the smartphone and location sensors technology and the prevalence of using location-based social networks alongside the improvements in artificial intelligence and machine learning techniques provide an excellent opportunity to exploit massive amounts of historical and real-time contextual information to recognise mobility patterns and achieve more accurate and intelligent predictions. This survey provides a comprehensive overview of the next useful location prediction problem with context-awareness. First, we explain the concepts of context and context-awareness and define the next location prediction problem. Then we analyse nearly thirty studies in this field concerning the prediction method, the challenges addressed, the datasets and metrics used for training and evaluating the model, and the types of context incorporated. Finally, we discuss the advantages and disadvantages of different approaches, focusing on the usefulness of the predicted location and identifying the open challenges and future work on this subject by introducing two potential use cases of next location prediction in the automotive industry.
\end{abstract}


\section{Introduction}
\label{introduction}

The prediction of mobility patterns holds significant importance across a spectrum of applications, ranging from urban planning and intelligent transportation systems to resource management in personalized medical services, tailored recommendation systems, and mobile communications. Anticipating mobility trends equips decision-makers with the tools to address challenges like traffic congestion by formulating realistic transportation plans and efficient scheduling strategies. Online ride-hailing and ride-sharing platforms rely heavily on precise mobility predictions to gauge travel demands, enabling more streamlined resource planning. Additionally, these predictions facilitate the identification of users with similar destinations, enhancing match-making, optimizing scheduling strategies, and ultimately reducing operational costs and energy consumption.

The advent of ubiquitous computing, coupled with advancements in Internet of Things (IoT), sensor advancements, and the extensive use of cellular networks, smartphones, and GPS gadgets, has generated vast repositories of mobility data collected from diverse sources. Simultaneously, the rapid growth of social networks such as Twitter, Facebook, and Foursquare has given rise to spatiotemporal mobility data enriched with contextual information such as text, images, videos, and user activity, preferences, and sentiments. The presence of abundant mobility data, coupled with the capabilities of artificial intelligence (AI) and machine learning (ML) methods, offers an unparalleled chance for researchers to tackle the task of predicting mobility. This involves harnessing diverse contextual information from various sources to construct robust models, leading to enhanced predictive accuracy.

In the realm of mobility prediction, an essential and valuable aspect is the anticipation of an individual's upcoming locations or destinations. This form of prediction finds applicability across diverse domains, including public health monitoring \cite{barlacchi2017you, canzian2015trajectories, pappalardo2016analytical}, traffic congestion management \cite{shi2019survey}, location-sensitive advertising enhancement, and link prediction within social network platforms \cite{wang2011human, zhu2015modeling}. Forecasting an individual's forthcoming location(s) is inherently challenging, as it necessitates capturing their habitual spatiotemporal mobility patterns, discerning their intention to explore new venues at different times, and identifying potential new places they may consider visiting. While individuals often adhere to regular movement patterns, an element of randomness in their mobility behavior complicates precise prediction of their whereabouts \cite{song2010limits}. Moreover, the integration of contextual information that influences human decisions regarding destination selection remains an ongoing challenge. Researchers have endeavored to harness various types of contextual information in the prediction process, including spatial context (e.g., historical trajectories), temporal context (e.g., day of the week, hour of the day, both for past visits and predictions), social ties (e.g., information related to individuals in social relationships), user profiles (e.g., preferences, occupation, schedule), and environmental context (e.g., spatial maps, details of visited and unvisited venues in the vicinity).

Numerous survey papers have comprehensively reviewed studies centered around mobility \cite{Wu2018, Luca2021, Zheng2015, Xu2020, Feng2016, Barbosa2018}. However, two characteristics are common across these reviews. Firstly, the majority of these reviews emphasise the methodologies employed for capturing mobility patterns or select studies based on their methodologies (e.g., \cite{Luca2021} examines studies employing deep learning-based approaches for human mobility analysis). Secondly, a prevailing tendency among these surveys is to treat the task of predicting the next location as synonymous with trajectory prediction \cite{Wu2018, Zheng2015}, thereby implying a primary reliance on spatiotemporal context—specifically location and time. However, this presumption is arguable. While spatiotemporal context is indeed pivotal for forecasting future locations, the potential contributions of other contextual factors to model efficacy should not be overlooked.

This paper serves as a comprehensive synthesis of context-aware next location prediction over the past decade. It centers on the paramount significance of leveraging diverse contextual information and the methods to seamlessly integrate it into the prediction process. The focal point lies in the examination of state-of-the-art technologies and research accomplishments from different perspectives, encompassing methodologies employed, challenges addressed, types of context employed, datasets utilized for training and metrics for evaluation. The major contributions of this paper can be encapsulated as follows:

\begin{itemize}
\item{We present an overview of the next location prediction problem from the perspective of context-awareness. To the best of our knowledge, this survey is the first survey examining the next location prediction problem from this point of view.}
\item{We classify the next location prediction methods into five main categories and introduce their characteristics and research progress. We also discuss the ability of these categories of methods to incorporate different types of contextual information.}
\item{We classify different types of datasets utilised for training and evaluating the next location prediction methods and examine their strengths and weaknesses.}
\item{We survey nearly thirty recent studies that focus on the next location prediction task and describe their advantages and disadvantages from the perspective of context-awareness and prediction performance.}
\item{We present two use cases of the next location prediction (as future work in this field) in the automotive cybersecurity sector to highlight the importance of the concept of \emph{useful next location prediction} and demonstrate the significance of employing different types of contextual information in the prediction task.}
\item{We highlight several open challenges and point out possible future research directions.}
\end{itemize}

The remainder of this paper is organised as follows: In Section \ref{preliminaries}, we lay the foundational concepts of context and context-awareness, clarify the next location prediction problem, and present the notion of context-aware next location prediction. Additionally, we explore the array of context sources that furnish invaluable insights and enrich the predictive task. Section \ref{prediction_methods} classifies the research endeavors in this domain according to their prediction methodologies, analysing them from varying vantage points such as context-awareness and the diversity of incorporated context types. Section \ref{prediction_methods} categorises the research studies conducted in this field based on their prediction method and analyses them from different perspectives, such as context-awareness and context types incorporated into the prediction task. In section \ref{datasets_and_evaluation_metrics}, we introduce different datasets commonly used by the studies in this field for training and evaluating the models and discuss the evaluation metrics used by them in order to assess the performance of the model. Section \ref{discussion} discusses the advantages and disadvantages of the studies analysed in the previous section. In section \ref{future_work}, we explain the open challenges and the future workarounds and finally, we conclude this survey in section \ref{conclusion}.

\section{Preliminaries}
\label{preliminaries}
\subsection{Context and context-awareness}
\label{context_and_contextawareness}
Context constitutes a fundamental attribute within contemporary IoT-enabled systems. Over recent decades, numerous scholars have endeavored to conceptually and operationally delineate context \cite{brown1995stick, hull1997towards, dey2001understanding}. The genesis of context-awareness is credited to Schilit and Theimer \cite{schilit1994disseminating}. In essence, context-awareness encompasses software that "adapts according to its location of use, the collection of nearby people and objects as well as changes to those objects overtime". This trait characterizes software's responsiveness to its environment. Dey's interpretation, which defines context as "any information that can be used to characterise the situation of an entity" has gathered widespread acceptance. Put differently, any data generated or consumed by a system is an intrinsic part of its contextual makeup. Stated simply, a system can be deemed context-aware if it leverages contextual insights to enhance its performance, efficiency, effectiveness, and overall utility.

Context exhibits varying levels of abstraction. For instance, consider temperature inference based on raw data from a thermometer in binary format, denoted in Centigrade/Fahrenheit units (e.g., 24\degree C / 75.2\degree F), and the subsequent categorization of weather warmth (e.g., hot/warm/cool/cold) stemming from this data. Both instances embody context, albeit at distinct abstraction tiers. However, the level of abstraction in context integrated into operations, such as prediction, can wield an impact on performance. Sigg \cite{sigg2008development} introduces an alternative, computation-centric stance, classifying the degree of abstraction in context based on the extent of preprocessing applied to raw data. While this approach introduces reasoned differentiation, it fails to delineate between high-level context and the notion of situation. Padovitz et al. \cite{padovitz12006unifying} present a three-tier hierarchy of abstraction levels termed the Context-Situation pyramid, aimed at addressing this limitation. As per their delineation, the most rudimentary data form is "sensory-originated data," employed, perhaps in conjunction with computation, to construct a contextual notion. In this schema, "context" assumes the role of information employed within a model to simulate real-world scenarios. On a higher plane, "situation" emerges as a meta-level concept, inferred through contextual analysis.
        
Within this survey, for the purpose of distinguishing between context-aware and non-context-aware methodologies, we adhere to the delineations and classifications presented by Padovitz et al. \cite{padovitz12006unifying} and Sigg \cite{sigg2008development}. Our criterion for regarding information as context necessitates its preliminary processing and partial or complete comprehensibility by humans. To clarify, consider instances such as 37.811760,144.964821 or 37\degree 48'42.3"S 144\degree57'53.4"E; these represent raw sensory-originated data. In contrast, designations like "McDonald's" encapsulate context, while descriptors such as "eating" or "working" delineate the situation. Consequently, we adopt the view that a model qualifies as context-aware if it assimilates contextual elements into its operational framework.

\subsection{Context-Aware Next Location Prediction}
{\bf Problem Definition.} Next location prediction involves foreseeing the subsequent location (stay point) that an individual or object will venture to, based on their historical mobility data. In a formal context, let \(u\) represent a user, \(T_u\) their trajectory, and \(p_t \in T_u\) their present location. The primary objective is to predict \(u\)'s forthcoming destination \(p_{t+1}\). This challenge can be approached through two distinct paradigms: (i) as a multi-class classification task, wherein the number of classes corresponds to the various locations, aiming to predict the next possible venue to be visited \(p_{t+1}\); (ii) as a regression task, projecting \(p_{t+1}\) = (\(x_{t+1}\), \(y_{t+1}\)), whereby \(x_{t+1}\) and \(y_{t+1}\) denote the geographical coordinates of the next location.

{\bf Next Point of Interest (POI) Recommendation:} Recommending Points of Interest (POIs) constitutes a pivotal facet of location-based social networks (LBSNs). This task aims to recommend POIs that the user may be interested in visiting in the future. Many studies addressed this problem, and one of the primary and popular approaches to it (like any other recommendation problem) is collaborative filtering, and its variants \cite{yu2015survey}. While the general task of POI recommendation may not directly correlate with the next location prediction, a variant of it is. Preceding research within the domain of next location prediction has underscored that user mobility entails a blend of established movement patterns (regularity) and ventures into novel or less-frequented locales (exploring) \cite{lian_ceprcollaborative_2015}. The latter is where the next location prediction problem meets the POI recommendation task.

A variant of POI recommendation endeavors to forecast the subsequent (sequential/successive) POI that an individual is poised to visit \cite{zhao2016stellar,zhang2020next}. This variant can also be considered as a variant of the next location prediction problem with an emphasis on the "exploring" facet. Another distinguishing characteristic lies in the fact that the next location recommendation task predominantly considers users' interests, while the next location prediction problem delves into users' intentions. Irrespective of the precise delineation, next location predictors yield a ranking that signifies the likelihood of each location serving as \(u\)'s next destination during their journey. 
% Notably, ascertaining advantageous locations for vehicles within our context of cybersecurity applications (OTA updates and window of opportunity) necessitates addressing not solely novel POIs but also habitual mobility patterns. 
In this paper, we will focus on the main problem of next location prediction, and we will exclude the studies only focusing on next POI recommendation.\\

Many attempts were made to address the mobility, specifically, the next location prediction problem \cite{cho_friendship_2011, liu_predicting_2016, feng_deepmove_2018, monreale_wherenext_2009, ying_semantic_2011, gambs_next_2012}. However, most of the techniques proposed by the studies are based on the historical movement of a user/object, trying to extract the behaviour and movement patterns to predict possible future locations. These methods usually have a training phase to extract regular movement patterns to predict future locations. While human mobility patterns typically adhere to regular spatial-temporal rhythms \cite{song2010limits}—particularly evident in urban settings, where individuals predominantly navigate around their dwellings and workplaces—there exist instances of infrequent visits to specific locales (e.g. shopping centers, museums, friends' residences) and even sporadic deviations from established routines. Besides that, people occasionally change their routines and try to visit and explore new venues (e.g. a restaurant recommended by a location-based social network) or meet some friends in a bar chosen by them. Hence, human mobility is evidently steered by more than mere regularity, exhibiting significant uncertainty influenced by lateral unpredictability. This inherent randomness and uncertainty limit the predictability of mobility. However, it is still not impossible to overcome this uncertainty if the information and context for tackling this issue are chosen carefully. 

This lateral randomness brings us to the exigency of context-aware mobility prediction, where not only a user's movement patterns should be extracted from the historical behaviour but, more importantly, the information characterising the situation of the user, his/her environment and any other entity in interaction with him/her, should be exploited to predict his/her possible future location. For example, a meeting scheduled in a user's diary, which will take place in a couple of hours in a restaurant, is valuable information in predicting his/her future whereabouts, while neglecting it may lead to a false prediction. Considering sunny weather on the weekend can help predict the user will probably go to a beach where not taking the weather into account may not result in the same outcome. Furthermore, consider the case of a college student whose movement patterns harmonize with class locations throughout an entire semester. As the subsequent semester starts, the student's schedule may experience a complete upheaval. Herein, a model founded solely on historical movement trends might necessitate the entirety of the semester to assimilate this new information accurately. Conversely, a context-aware model that integrates the student's university semester schedule is poised to yield enhanced precision in predictions.

\subsection{Context Categories}
Although the definitions of context presented in section \ref{context_and_contextawareness} give us an intuition about the concept of context, situation and context-awareness, for operational use of context, such definitions can be general and incomplete, leading to weak practical usefulness. In a bid to introduce a more actionable delineation of context, Zimmermann et al. \cite{zimmermann2007operational} provide an operational framework wherein context is classified into five general categories: individual, location, time, activity, and relational. We use this categorical schema summarised below, to identify the contextual information incorporated to the next location prediction task.

\begin{itemize}
\item {\bf Individual Context:} Encompasses contextual information about an entity or a collective of entities to which the context pertains. This entity could be human, natural, or artificial.
\item {\bf Location Context:} Encompasses information that delineates the spatial coordinates of an entity.
\item {\bf Time Context:} Encompasses information pertaining to the temporal dimension, encompassing variables like time of day, day of the week, season, and current time.
\item {\bf Activity Context:} Encompasses details about the activities undertaken by entities, spanning actions such as shopping, dining, or sleeping, and encompassing activities the entity is presently engaged in or potentially involved in the future.
\item {\bf Relational Context:} Pertains to the interconnections between an entity and other entities, encapsulating aspects such as social ties in online networks, relationships within families, and entities that share physical connections.
\end{itemize}

% ========================== III. Prediction Methods ===========% 
% ==============================================================% 
\section{Prediction Methods}
\label{prediction_methods}
This section delves into an in-depth discussion and categorisation of the various research endeavors centered around the next location prediction task. The analyses encompass their context-awareness levels, employed methodologies, and integrated context types. From a comprehensive pool of closely correlated research papers, twenty-nine pertinent studies were chosen and analysed with respect to the aforementioned facets. 

In the quest for these studies, an exhaustive bibliographic exploration was undertaken across IEEE, ACM, ScienceDirect, and Springer. This selection was based on the quality and pertinence of the publications in relation to the subject. Employing a purposeful search string ("location prediction" OR "next place prediction" OR "destination prediction" OR "location"), some filters was applied to the paper selection process:

\begin{enumerate}
\item \emph{Publication Date:} The papers were confined to those published within the temporal span ranging from 2012 to 2022. A couple of exceptions were admitted due to their importance in the domain.
\item \emph{Publication Type and Citations:} Papers that were presented in workshops, symposiums, or cited less than 100 times were not considered. This rule included recent papers that gained attention.
\item \emph{Relevance:}We carefully reviewed the abstracts of each paper to determine their relevance to the context, resulting in a final selection of twenty-nine articles. It's worth noting that surveys related to next location prediction were not included in this group.
\end{enumerate}

As a summary, Table \ref{tab:table1} provides an overview of the selected papers concerning challenges they addressed, methods exploited, datasets used for training/evaluation, evaluation metrics and the context types they incorporated in their model.

\subsection{Evidence-based methods}
A significant theoretical approach called the theory of belief functions, also known as evidence theory or Dempster–Shafer theory (DST), provides a flexible framework for handling uncertainty. This theory has connections to other frameworks like probability, possibility, and imprecise probability theories \cite{gordon1984dempster}. Dempster-Shafer theory (DST) has gained attention for its effective approach in addressing challenges related to combining different pieces of evidence to make decisions in situations with high uncertainty. One key advantage of this theory is its ability to model the refinement of hypotheses through accumulating evidence, while explicitly representing uncertainty due to lack of knowledge or reservations in judgment. The theory is based on two main principles: assigning belief degrees to hypotheses and using Dempster's rule to combine these degrees from different bodies of evidence. The core of Dempster's rule lies in its ability to support consistent evidence while handling conflicting information.

Illustratively, Samaan et al. \cite{samaan_mobility_2005} introduced a predictive framework based on Dempster-Shafer theory. This innovative effort marked the initial attempt to tackle the next location prediction challenge by incorporating context. The framework orchestrates the integration and reasoning of contextual information, including real-world maps, user profiles, preferences, tasks, and schedules. Leveraging Dempster-Shafer theory, the model aims to understand the user's behavioral patterns in relation to their decisions about upcoming locations. This approach enables the model to adapt to new scenarios and enhance its effectiveness by combining contextual hints, allowing it to predict the user's mobility trajectory and future destination with substantial confidence.

\subsection{Probabilistic and Distribution-based Methods}
Methods that fall into the probabilistic and distribution-based group usually attempt to model human mobility using two approaches. First, modelling mobility patterns using distributions of location and time (e.g. Gaussian or Gaussian mixture models). Second, computing the plausibility of a user arriving at a location using probabilistic models (e.g. Bayes theorem) \cite{cho_friendship_2011, do_contextual_2012, gao_exploring_2012, gao_modeling_2013}. 

Cho and his colleagues \cite{cho_friendship_2011} explored how individuals travel and connect with others in social networks. They investigated how our daily habits and friendships impact the way we go from one place to another. Through studying data about people's movements, they discovered some fascinating trends in our behavior. They found that short trips tend to follow a consistent pattern in specific locations and times, regardless of our friends. However, when we travel longer distances, our social relationships become more significant. Surprisingly, our social connections can explain about 10 to 30 per cent of how we move, while regular behaviors in certain places account for around 50 to 70 per cent of the patterns in how we move around.

The authors introduce a model called the Periodic Mobility Model (PMM), which is a comprehensive framework consisting of two main elements: the places a person regularly visits and the shifts between these places over time. This model effectively captures the complex patterns of how someone moves between different locations using a model that considers the specific patterns for each day. The central idea of the PMM is how it represents places in a detailed way. It achieves this by using a combination of Gaussian models centered around two key places: "home" and "work." To explain further, the authors create a probability distribution that describes the person's state (\(P[c_u(t)]\)) by using a specific type of Gaussian distribution that is based on time:

\begin{equation}
    N_{(H/W)}(t) = \frac{P_{c(H/W)}}{\sqrt{2\pi\sigma^2_{(H/W)}}}\:exp\left[-\left(\frac{\pi}{12}\right)^2\frac{(t-\tau_{(H/W)})^2}{2\sigma^2_{(H/W)}}\right]
\end{equation}

Subsequently, the probability of the user's state \(c_u(t)\) being either "home" or "work" is determined by the ratio of the Gaussian models, leading to the following expression:

\begin{equation}
    P[c_u(t)=(H \text{ or } W)] = \frac{N_{H/W}(t)}{N_{H/W}(t)+N_{W/H}(t)}
\end{equation}

Upon establishing the user's state at a given time, the authors proceed to compute the likelihood of the user's next location, which hinges upon the amalgamation of the "home" and "work" distributions. This computation is framed as follows:

\begin{equation}
    P[x(t)=x] = P[x_u(t)=x|c_u(t)=H] \cdot P[c_u(t)=H] + P[x_u(t)=x|c_u(t)=W] \cdot P[c_u(t)=W]
\end{equation}

Here, \(\tau_{(H/W)}\) signifies the average time of day when a user typically assumes the "home" or "work" state, \(\sigma_{(H/W)}\) corresponds to the variance associated with the time of day for these respective states, and \(P_{c{(H/W)}}\) represents the time-independent probability linked to the generation of any given check-in within the "home" or "work" state.

This study can be categorized as partially context-aware, as it harnesses low-level contextual information to facilitate the prediction of the subsequent location. These context-rich elements encompass the day of the week, the hour of the day, the semantic implications of the GPS traces (such as "home" and "work"), and the presence of interpersonal affiliations (such as friendship) among users.

\begin{figure}
    \centering
    \includegraphics[width=3in]{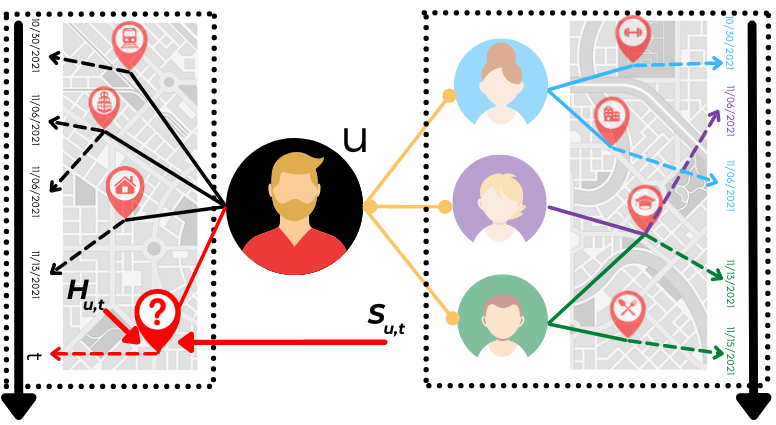}
    \caption{Contextual information involved in modeling user u's mobility behaviour \cite{gao_modeling_2013}}
    \label{SHM-T}
\end{figure}

Similarly, Gao et al. \cite{gao_modeling_2013} adopt a probabilistic and distribution-based methodology for addressing this task. They assume that location-based social network data comprise three distinct layers of information: a social layer, a geographical layer, and a temporal layer. Their investigation hones in on the temporal layer, delving into the ramifications of temporal dynamics and cyclic patterns within human mobility. Illustrated in Figure \ref{SHM-T}, this research discerns the configuration of contextual spatiotemporal and social information designed to model the behavior of user \(u\). Here, \(H_{u,t}\) and \(S_{u,t}\) signify the observed historical check-in actions of user \(u\) and their friends respectively, leading up to time \(t\).
    
Various temporal indications linked to cyclic patterns can be inferred from \(t\), signifying a user's check-in state, encompassing factors like the hour of the day, day of the week, month of the year, and more. Given the corresponding observations of \(H_{u,t}\) and \(S_{u,t}\), along with the temporal context represented by \(r(t)\), the probability distribution over user \(u\)'s check-in locations at time \(t\) is defined in Equation \ref{SHM-T-eq1}, where \(c_u\) denotes the user's current location and \(l\) represents an arbitrary location.

This probability distribution is disassembled into spatial and temporal components by applying Bayes's theorem (Equation \ref{SHM-T-eq1}). The spatial component is computed using the Social-Historical Model (SHM) proposed by Gao et al. \cite{gao_exploring_2012}, which models mobility behavior using the Pitman-Yor Process. The Pitman-Yor Process is a stochastic process, with its sample path forming a probability distribution.
\begin{equation}
\label{SHM-T-eq1}
    P(c_u=l|r(t),H_{u,t},s_{u,t}) \propto P(r(t)|c_u=l,H_{u,t},s_{u,t})P(c_u=l|H_{u,t},s_{u,t})
\end{equation}
The second component, the temporal component, is formulated as a probability function that encapsulates the combined effects of personal preferences from \(H_{u,t}\) and social influences from \(S_{u,t}\).
\begin{equation}
    P(r(t)|c_u=l,H_{u,t},s_{u,t}) \propto
    \alpha P(r(t)|c_u=l,H_{u,t}) + (1-\alpha)P(r(t)|c_u=l,S_{u,t})=
\end{equation}
\begin{equation*}
    \alpha P(r(t)|c_u=l,H_{u,t})+
    (1-\alpha)\frac{\sum_{u_i\in F(u)}sim(u,u_i)P(r(t)|c_u=l,H_{u,t})}{\sum_{u_i\in F(u)}sim(u,u_i)}
\end{equation*}
Where \(\alpha\) is a parameter governing the contribution of personal preferences and social influence, \(F(u)\) represents the group of user \(u\)'s social friends, and \(sim(u,u_i)\) stands for the cosine similarity between the visited locations of \(u\) and \(u_i\).

\subsection{Deep neural networks-based methods}
Recently, deep learning models such as Recurrent Neural Networks (RNNs) and Convolutional Neural Networks (CNNs) have been applied to a range of spatiotemporal problems, including traffic flow prediction \cite{lv2014traffic, polson2017deep}, traffic congestion prediction \cite{ma2015large}, on-demand services \cite{wang2017deepsd}, and successive POI recommendation \cite{huang_attention-based_2021}. The task of next location prediction, being a significant sub-task within spatiotemporal problems, has also been addressed using deep neural networks by various studies \cite{liu_predicting_2016, feng_deepmove_2018, yao_serm_2017, kong_hst-lstm_2018, chen_context-aware_2020, yang_location_2020, feng_pmf_2020}. These models have demonstrated remarkable performance owing to their inherent capabilities of automatic feature representation, selection learning, and function approximation. Nonetheless, these models are not without limitations. Primarily, they require substantial amounts of training data. Additionally, their interpretability is often lower compared to traditional machine learning approaches.

Liu et al. \cite{liu_predicting_2016} propose an architecture based on Recurrent Neural Networks (RNNs), introducing two modifications to the transitional matrices within the RNN structure. They integrate time-specific and distance-specific transitional matrices into the original matrix to capture the influence of elements from the most recent history and the variations in geographical distances between locations, respectively.

Conversely, Feng et al. \cite{feng_deepmove_2018} present an attention-based recurrent model called DeepMove for predicting human mobility from extensive yet sparse trajectories. DeepMove addresses three primary challenges: I) Capturing the intricate sequential transition patterns characterized by their time-dependent and high-order nature. II) Accounting for the multi-level periodicity inherent in human mobility. III) Handling the heterogeneity and sparsity present in the collected trajectory data.

Incorporating even more contextual information into the prediction task, Sun et al. \cite{sun2022predicting} endeavor to address the next location prediction problem by employing semantic time-series and leveraging Recurrent Neural Networks (RNNs). They argue that the separation of semantic knowledge from spatiotemporal data impedes the understanding of user activities and the relationships between users and places. As discussed in Section \ref{Datasets}, check-in data provides semantic insights about places. However, it often suffers from data sparsity issues due to voluntary user data recording, resulting in infrequent data records separated by hours or even days. To mitigate the sparsity challenge in Location-Based Social Network (LBSN) data, the authors identify that users' check-ins during various periods exhibit regularity, indicating fixed behavioral patterns during specific time periods. This insight leads them to fill in missing data for these periods. Fragmented period data is then interpolated using the Continuous Bag-of-Words (CBOW) technique. CBOW, a common method in natural language processing, predicts a target word based on its context. Building upon Cho et al.'s \cite{cho_friendship_2011} findings, which indicate that friendship explains around 5\% of user behavior while the effect of users with similar behavior patterns accounts for 40\%, Sun et al. introduce the concept of "virtual friends." Virtual friends are users who share comparable behavior patterns with a given user, even if they are not acquainted in real life. The concept of virtual friends, coupled with the integration of behavioral similarity, proves highly effective in mitigating the cold start problem.

\subsection{Pattern-based methods}
An effective approach for addressing the next location prediction challenge involves utilizing a subset of data and pattern mining techniques, commonly known as trajectory mining. This method involves mining users' trajectories, which consist of sequences of locations and corresponding timestamps, to identify frequent sequential patterns that can be leveraged for predicting future locations.

A pioneering study in this category is undertaken by Monreale et al. \cite{monreale_wherenext_2009}. This study focuses on extracting frequent movement patterns, termed T-patterns, utilizing a trajectory pattern mining algorithm. These T-patterns are then aggregated to construct a prefix tree referred to as the T-pattern tree. In this tree, nodes represent regions that are frequently visited, while edges symbolize travel between these regions along with associated typical travel times. Common prefixes among T-patterns translate into shared paths within this tree. Ultimately, the T-pattern tree serves as the basis for predicting the future location of a moving entity. Notably, since this approach employs all trajectories irrespective of individual users, it implicitly addresses challenges associated with the cold start problem and data sparsity.

Ying et al. \cite{ying_semantic_2011} also harness raw spatiotemporal data, i.e., trajectories, for predicting subsequent locations. However, In terms of context-awareness, this study introduces semantic significance to clusters of locations (referred to as stay points) through the utilization of geographic semantic information databases and reverse geocoding. Their proposed location prediction framework, known as SemanPredict, strives to uncover semantic trajectory patterns of individuals and categorize users based on their mobility behavior, leveraging maximal semantic trajectory pattern similarity between different users. The final step involves using frequent semantic trajectory patterns from similar users to predict a user's upcoming stay point. As this framework solely considers the sequence of semantic stay points, temporal context does not play a central role in their proposed method.
    
\subsection{Markov model-based methods}
Markov chains, also known as Markov processes, are stochastic processes that describe a sequence of potential events while assuming the Markov property \cite{gagniuc2017markov}. These chains are frequently employed to model a user's location history by considering the transition probabilities between different locations. This approach assumes that the prediction of a user's next location relies on their current or recent locations. Markov-based models offer a popular means of capturing mobility patterns and forecasting the subsequent state.

Gambs et al. \cite{gambs_next_2012} propose a two-phase model for predicting the next location. The first phase involves extracting low-level context from raw spatiotemporal data to cluster trajectory data into meaningful stay places, such as "home," "work," and "unknown." This low-level contextual information encompasses semantic labels of the clusters, travel time between points, staying duration at a location, and speed for filtering on-the-move points. The second phase employs an n-Mobility Markov Chain (2-MMC), where states correspond not only to individual POIs but also encompass sequences of the two preceding visited POIs.

Another attempt to address the next location prediction challenge is pursued by Lv et al. \cite{lv_big_2017}. This study introduces a framework grounded in Markov chains. Their innovative contribution lies in devising distinct models for various mobility behaviors, tailored to individual users' living habits. The process begins with the extraction of Points of Interest (POIs) from trajectories via a clustering algorithm. Subsequently, these identified POIs are used to transform point-based trajectories into travel sequences. In the end, users are sorted into four categories, namely "Day Postman," "Family Person," "Party Person," and "Hard Postman," based on the entropy in their travel behavior.

\begin{figure}
    \centering
    \includegraphics[width=3.5in]{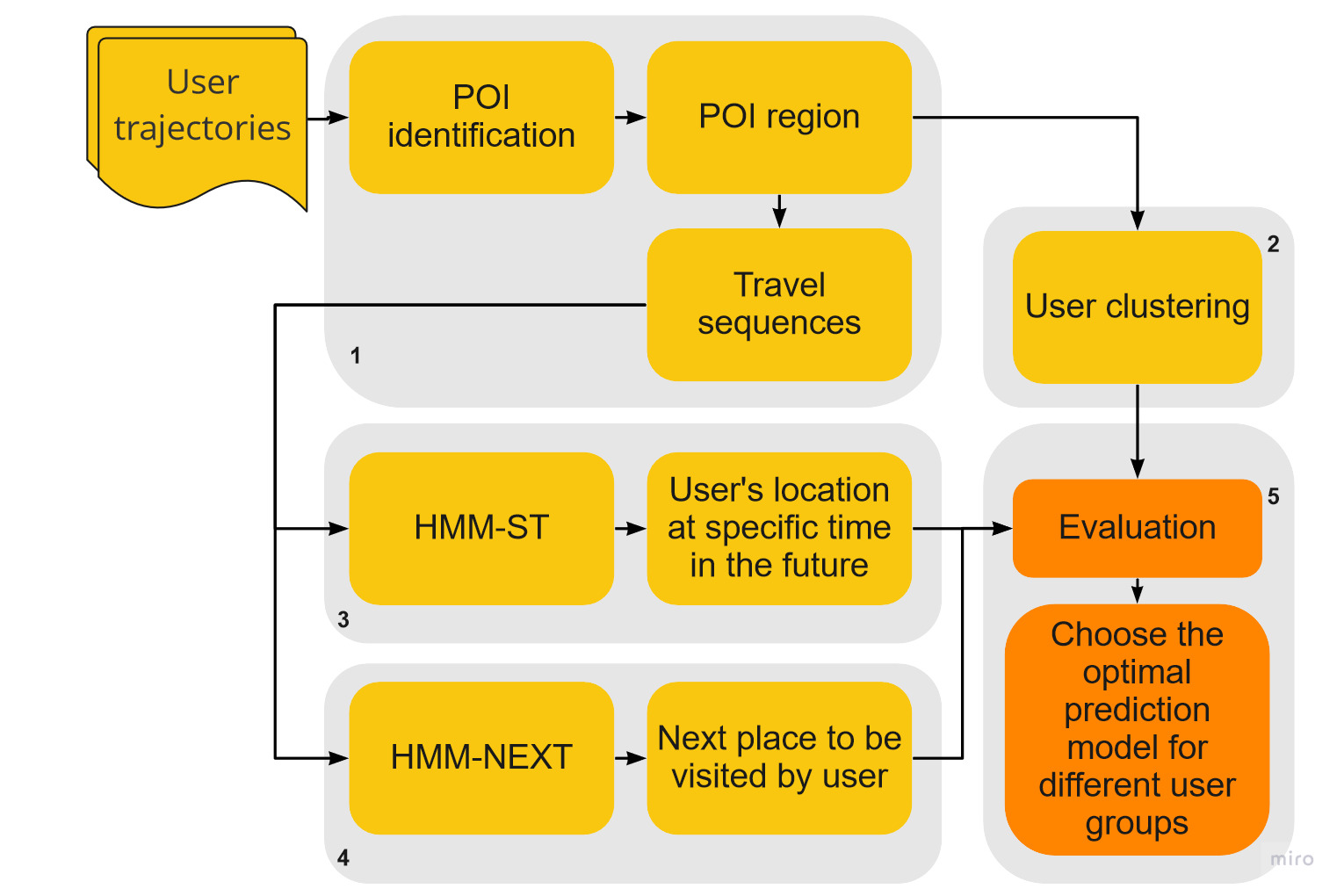}
    \caption{Structure of the prediction framework proposed by \cite{lv_big_2017}.}
    \label{HMM-STandNEXT}
\end{figure}

Enhanced Markov models are introduced for prediction purposes. The first model, termed HMM-Based Spatiotemporal Prediction (HMM-ST), forecasts a user's location at a specific future time. The second model, known as HMM-Based Next-Place Prediction (HMM-NEXT), predicts a user's subsequent location upon leaving their current place. By comparing the performance of these models across user categories, the study suggests incorporating users' living habits to enhance model performance and flexibly applying mobility predictors based on individual models. The proposed structure of this model is illustrated in Fig. \ref{HMM-STandNEXT}, where Stages 1-5 correspond to the previously outlined procedures.
\subsection{Hybrid methods}
Hybrid methods for mobility prediction leverage diverse techniques to model different aspects of the problem. Wang et al. \cite{wang_regularity_2015} adopt a hybrid approach by integrating regularity and conformity aspects of human mobility into a unified model. This study introduces the RCH (Regularity, Conformity, Heterogeneous) hybrid predictive model, which combines both the regularity (characterizing the consistent nature of human mobility) and conformity (reflecting how people's movements are influenced by others) aspects of human mobility. Regularity is modeled using the gravity model, while conformity is addressed using collaborative filtering. To boost predictive accuracy, the model acquires location profiles from diverse mobility datasets using a gravity model.

For predicting a user \(u_i\)'s visit to venue \(v_j\), two key components are considered: regularity (\(R_{ij}^{(r)}(t)\)) and conformity (\(R_{ij}^{(c)}(t)\)). The regularity term is defined by integrating the visiting frequency of a grid cell and the transition probability between different grid cells, influenced by the gravity model. The conformity term is modeled based on the social conformity theory, considering users' similarities in terms of backgrounds, interests, and social status. The hybrid model effectively combines these components to predict users' mobility patterns and locations.

\begin{landscape}
\thispagestyle{empty}
\newgeometry{vmargin=180pt, hmargin=30pt}

\begin{table*}
\small
    \centering
\caption{Summary of Next Location Prediction papers (ordered chronologically)}
    \begin{center}
\label{tab:table1}
\begin{tabular}{>{\centering}p{2cm}>{\centering}p{0.5cm}>{\centering}p{5cm}>{\centering}p{3cm}>{\centering}p{2cm}>{\centering}p{2cm}>{\centering}p{1.5cm}>{\centering}p{.1cm}>{\centering}p{.1cm}>{\centering}p{.1cm}>{\centering}p{.1cm}>{\centering}p{.1cm}>{\centering}p{2cm}}
\hline
\multirow{2}{*}{Authors} & \multirow{2}{*}{Year} & \multirow{2}{*}{Challenges addressed} & \multirow{2}{*}{Model} & \multirow{2}{*}{Method} & \multirow{2}{*}{Datasets} & \multirow{2}{*}{Metrics} & \multicolumn{6}{c}{Context types} \\ \cline{8-13} 
 & & & & & & & \multicolumn{1}{c}{\rotatebox[origin=c]{90}{Loc.}} & \multicolumn{1}{c}{\rotatebox[origin=c]{90}{Time}} & \multicolumn{1}{c}{\rotatebox[origin=c]{90}{Rel.}} & \multicolumn{1}{c}{\rotatebox[origin=c]{90}{Act.}} & \multicolumn{1}{c}{\rotatebox[origin=c]{90}{Ind.}} & \multicolumn{1}{c}{Other} \\
\hline
Samaan et al. \cite{samaan_mobility_2005} & 2005 & cold start - randomness of human mobility & - * & Evidence theory-based & self collected GPS data - questionnaires & Accuracy & \multicolumn{1}{c|}{\checkmark} & \multicolumn{1}{c|}{\checkmark} & \multicolumn{1}{c|}{} & \multicolumn{1}{c|}{\checkmark} & \multicolumn{1}{c|}{\checkmark} & spatial maps containing semantic information \\ \cdashline{1-7} \cline{8-13}
Monreale et al. \cite{monreale_wherenext_2009} & 2009 & cold start - data sparsity - Accuracy & WhereNext & Pattern-based & GeoPKDD & Coverage - Accuracy & \multicolumn{1}{c|}{\checkmark} & \multicolumn{1}{c|}{\checkmark} & \multicolumn{1}{c|}{} & \multicolumn{1}{c|}{} & \multicolumn{1}{c|}{} & \\ \cdashline{1-7} \cline{8-13}
Cho et al. \cite{cho_friendship_2011} & 2011 & Examine the influence of social routines on patterns of human mobility, as well as the impact of social connections, such as friends whom individuals travel to meet & Periodic Mobility Model (PMM) and Periodic \& Social Mobility Model (PSMM) & Distribution-based & Gowalla - Brightkite - CDR & Accuracy - MAE & \multicolumn{1}{c|}{\checkmark} & \multicolumn{1}{c|}{\checkmark} & \multicolumn{1}{c|}{\checkmark} & \multicolumn{1}{c|}{} & \multicolumn{1}{c|}{} & \\ \cdashline{1-7} \cline{8-13} 
Ying et al. \cite{ying_semantic_2011} & 2011 & adding extra context to prediction (e.g. activity) by using semantic labels of locations instead of using coordinates or label less clusters & SemanPredict & Pattern-based & MIT reality dataset & Precision - Recall - F-measure & \multicolumn{1}{c|}{\checkmark} & \multicolumn{1}{c|}{\checkmark} & \multicolumn{1}{c|}{} & \multicolumn{1}{c|}{} & \multicolumn{1}{c|}{} & \\ \cdashline{1-7} \cline{8-13}
Do and Gatica-Perez \cite{do_contextual_2012} & 2012 & finding relevant contextual features - integrating general and personal prediction models to be used in "cold start" situations & - * & Probabilistic-based & Self collected GPS data & Accuracy & \multicolumn{1}{c|}{\checkmark} & \multicolumn{1}{c|}{\checkmark} & \multicolumn{1}{c|}{} & \multicolumn{1}{c|}{} & \multicolumn{1}{c|}{} & \\ \cdashline{1-7} \cline{8-13}
Xue et al. \cite{xue_destination_2013} & 2012 & data sparsity - privacy - computational performance & Sub-Trajectory Synthesis (SubSyn) & Markov-based & T-drive & Coverage - MAE & \multicolumn{1}{c|}{\checkmark} & \multicolumn{1}{c|}{\checkmark} & \multicolumn{1}{c|}{} & \multicolumn{1}{c|}{} & \multicolumn{1}{c|}{} & \\ \cdashline{1-7} \cline{8-13}
Noulas et al. \cite{noulas_mining_2012} & 2012 & feature selection and extraction & - * & - ** & Foursquare & APR - Accuracy & \multicolumn{1}{c|}{\checkmark} & \multicolumn{1}{c|}{\checkmark} & \multicolumn{1}{c|}{\checkmark} & \multicolumn{1}{c|}{} & \multicolumn{1}{c|}{} & contextual information related to historical visits and temporal context \\ \cdashline{1-7} \cline{8-13}
Gao et al. \cite{gao_exploring_2012} & 2012 & investigate the effect of social-historical ties on users' check-in behavior & social-historical model (SHM) & Probabilistic-based & Foursquare & Accuracy & \multicolumn{1}{c|}{\checkmark} & \multicolumn{1}{c|}{\checkmark} & \multicolumn{1}{c|}{\checkmark} & \multicolumn{1}{c|}{} & \multicolumn{1}{c|}{} & \\ \cdashline{1-7} \cline{8-13}
Gambs et al. \cite{gambs_next_2012} & 2012 & incorporate more than one previous state of the user into the Markov model in order to increase prediction accuracy & n-MMC & Markov-based & self collected GPS data - Geolife & Accuracy & \multicolumn{1}{c|}{\checkmark} & \multicolumn{1}{c|}{\checkmark} & \multicolumn{1}{c|}{} & \multicolumn{1}{c|}{} & \multicolumn{1}{c|}{} & motion speed \\ \cdashline{1-7} \cline{8-13}
Gao et al. \cite{gao_modeling_2013} & 2013 & analyze the effect of temporal layer of source of context in LBSNs & social-historical + temporal model (SHM + T) & Probabilistic-based & Foursquare - Brightkite & Accuracy & \multicolumn{1}{c|}{\checkmark} & \multicolumn{1}{c|}{\checkmark} & \multicolumn{1}{c|}{\checkmark} & \multicolumn{1}{c|}{} & \multicolumn{1}{c|}{} & \\ \cdashline{1-7} \cline{8-13}
Ying et al. \cite{ying_mining_2013} & 2013 & incorporate the geographic/temporal/semantic triggered intentions in prediction - incorporate similarity between user GTS pattern trees into prediction & Geographic-Temporal-Semantic-based Location Prediction (GTS-LP) & Pattern-based & EveryTrail - Bikely & Accuracy - Coverage - F-measure & \multicolumn{1}{c|}{\checkmark} & \multicolumn{1}{c|}{\checkmark} & \multicolumn{1}{c|}{} & \multicolumn{1}{c|}{} & \multicolumn{1}{c|}{} & stay time in each venue and transition time between them \\ \cdashline{1-7} \cline{8-13}
\end{tabular}
\end{center}
\begin{enumerate}
   \item[*] No name is assigned to the proposed model
   \item[**] No model is proposed
\end{enumerate}
\end{table*}
 
\end{landscape}

\begin{landscape}
\thispagestyle{empty}
\newgeometry{vmargin=180pt, hmargin=30pt}

\begin{table*}
% % \linespread{1}
\small
\ContinuedFloat
    \begin{center}
\caption{Summary of Next Location Prediction papers - continued (ordered chronologically)}
\label{tab:table1}
\begin{tabular}{>{\centering}p{2cm}>{\centering}p{0.5cm}>{\centering}p{5cm}>{\centering}p{3cm}>{\centering}p{2cm}>{\centering}p{2cm}>{\centering}p{1.5cm}>{\centering}p{.1cm}>{\centering}p{.1cm}>{\centering}p{.1cm}>{\centering}p{.1cm}>{\centering}p{.1cm}>{\centering}p{2cm}}
\hline
\multirow{2}{*}{Authors} & \multirow{2}{*}{Year} & \multirow{2}{*}{Challenges addressed} & \multirow{2}{*}{Model} & \multirow{2}{*}{Method} & \multirow{2}{*}{Datasets} & \multirow{2}{*}{Metrics} & \multicolumn{6}{c}{Context types} \\ \cline{8-13} 
 & & & & & & & \multicolumn{1}{c}{\rotatebox[origin=c]{90}{Loc.}} & \multicolumn{1}{c}{\rotatebox[origin=c]{90}{Time}} & \multicolumn{1}{c}{\rotatebox[origin=c]{90}{Rel.}} & \multicolumn{1}{c}{\rotatebox[origin=c]{90}{Act.}} & \multicolumn{1}{c}{\rotatebox[origin=c]{90}{Ind.}} & \multicolumn{1}{c}{Other} \\
\hline
Preoţiuc-Pietro and Cohn \cite{preotiuc-pietro_mining_2013} & 2013 & show the importance of temporal patterns in mobility prediction & - * & - ** & Foursquare & Accuracy & \multicolumn{1}{c|}{\checkmark} & \multicolumn{1}{c|}{\checkmark} & \multicolumn{1}{c|}{} & \multicolumn{1}{c|}{\checkmark} & \multicolumn{1}{c|}{} & \\ \cdashline{1-7} \cline{8-13}
Lian et al. \cite{lian_ceprcollaborative_2015} & 2015 & data sparsity - predict exploration probability to incorporate location recommendation into prediction task & Collaborative Exploration and Periodically Returning Model (CEPR) & Markov-based & Gowalla - Jiepang & Accuracy & \multicolumn{1}{c|}{\checkmark} & \multicolumn{1}{c|}{\checkmark} & \multicolumn{1}{c|}{\checkmark} & \multicolumn{1}{c|}{} & \multicolumn{1}{c|}{} & user, spatial and temporal context extracted from historical check-ins (e.g. user and location entropy) \\ \cdashline{1-7} \cline{8-13}
Wang et al. \cite{wang_regularity_2015} & 2015 & randomness (incorporate both regularity and conformity) - data sparsity - heterogeneous data & RCH-Regularity, Conformity, Heterogeneous & Gravity model / Collaborative filtering & Sina Weibo - Taxis and Buses GPS & acc@K - APR & \multicolumn{1}{c|}{\checkmark} & \multicolumn{1}{c|}{\checkmark} & \multicolumn{1}{c|}{} & \multicolumn{1}{c|}{} & \multicolumn{1}{c|}{} & \\ \cdashline{1-7} \cline{8-13}
Liu et al. \cite{liu_predicting_2016} & 2016 & capture the impact of elements in the most recent history - continuous time interval and geographical distance problem which causes data sparsity problem in transition matrices & Spatial Temporal Recurrent Neural Networks (ST-RNN) & Neural Networks & Global Terrorism - Gowalla & recall@ {[}1,5,10{]} - F1-score@ {[}1,5,10{]} - MAP - AUC & \multicolumn{1}{c|}{\checkmark} & \multicolumn{1}{c|}{\checkmark} & \multicolumn{1}{c|}{} & \multicolumn{1}{c|}{} & \multicolumn{1}{c|}{} & \\ \cdashline{1-7} \cline{8-13}
Lv et al. \cite{lv_big_2017} & 2017 & randomness - Integrate users' lifestyle patterns into the prediction process & HMM-ST \& HMM-NEXT & Markov-based & CDR & Accuracy & \multicolumn{1}{c|}{\checkmark} & \multicolumn{1}{c|}{\checkmark} & \multicolumn{1}{c|}{} & \multicolumn{1}{c|}{} & \multicolumn{1}{c|}{} & user's mobility habits (day postman / family person / party person /hard person) \\ \cdashline{1-7} \cline{8-13}
Yao et al. \cite{yao_serm_2017} & 2017 & incorporate semantic trajectories texts containing information for location prediction instead of classic GPS trajectories & Semantic-Enriched Recurrent Model (SERM) & Neural Networks & Foursquare - Twitter & acc@ {[}1,5,10,20{]} - MAE & \multicolumn{1}{c|}{\checkmark} & \multicolumn{1}{c|}{\checkmark} & \multicolumn{1}{c|}{} & \multicolumn{1}{c|}{\checkmark} & \multicolumn{1}{c|}{} & \\ \cdashline{1-7} \cline{8-13}
Feng et al. \cite{feng_deepmove_2018} & 2018 & the intricate sequential transition patterns observed with time-dependent and high-order characteristics - the multi-level periodic nature of human mobility - the diversity and scarcity of the gathered trajectory data & DeepMove & Neural Networks & Foursquare - CDR & Accuracy & \multicolumn{1}{c|}{\checkmark} & \multicolumn{1}{c|}{\checkmark} & \multicolumn{1}{c|}{} & \multicolumn{1}{c|}{} & \multicolumn{1}{c|}{} & \\ \cdashline{1-7} \cline{8-13}
Kong and Wu \cite{kong_hst-lstm_2018} & 2018 & data sparsity & HST-LSTM & Neural Networks & Baidu & acc@ {[}1,5,10,20{]} - MRR & \multicolumn{1}{c|}{\checkmark} & \multicolumn{1}{c|}{\checkmark} & \multicolumn{1}{c|}{} & \multicolumn{1}{c|}{} & \multicolumn{1}{c|}{} & \\ \cdashline{1-7} \cline{8-13}
Yang et al. \cite{yang_revisiting_2019} & 2019 & automate feature learning using graph embedding & LBSN2Vec & Graph Embedding & Foursquare & acc@10 & \multicolumn{1}{c|}{\checkmark} & \multicolumn{1}{c|}{\checkmark} & \multicolumn{1}{c|}{\checkmark} & \multicolumn{1}{c|}{\checkmark} & \multicolumn{1}{c|}{} & \\ \cdashline{1-7} \cline{8-13}
\end{tabular}
\end{center}
\begin{enumerate}
   \item[*] No name is assigned to the proposed model
   \item[**] No model is proposed
\end{enumerate}
\end{table*}
 
\end{landscape}

\begin{landscape}
\thispagestyle{empty}
\newgeometry{vmargin=180pt, hmargin=30pt}

\begin{table*}
% \linespread{1}
\small
    \ContinuedFloat
    \begin{center}
\caption{Summary of Next Location Prediction papers - continued (ordered chronologically)}
\label{tab:table1}
\begin{tabular}{>{\centering}p{2cm}>{\centering}p{0.5cm}>{\centering}p{5cm}>{\centering}p{3cm}>{\centering}p{2cm}>{\centering}p{2cm}>{\centering}p{1.5cm}>{\centering}p{.1cm}>{\centering}p{.1cm}>{\centering}p{.1cm}>{\centering}p{.1cm}>{\centering}p{.1cm}>{\centering}p{2cm}}
\hline
\multirow{2}{*}{Authors} & \multirow{2}{*}{Year} & \multirow{2}{*}{Challenges addressed} & \multirow{2}{*}{Model} & \multirow{2}{*}{Method} & \multirow{2}{*}{Datasets} & \multirow{2}{*}{Metrics} & \multicolumn{6}{c}{Context types} \\ \cline{8-13} 
 & & & & & & & \multicolumn{1}{c}{\rotatebox[origin=c]{90}{Loc.}} & \multicolumn{1}{c}{\rotatebox[origin=c]{90}{Time}} & \multicolumn{1}{c}{\rotatebox[origin=c]{90}{Rel.}} & \multicolumn{1}{c}{\rotatebox[origin=c]{90}{Act.}} & \multicolumn{1}{c}{\rotatebox[origin=c]{90}{Ind.}} & \multicolumn{1}{c}{Other} \\
\hline
Feng et al. \cite{feng_pmf_2020} & 2020 & Privacy -trade of between privacy preserving and performance/accuracy & Privacy-preserving Mobility prediction framework via Federated learning (PMF) & Neural Networks & Foursquare - Twitter - Self collected GPS data & acc@ {[}1,3,5{]} & \multicolumn{1}{c|}{\checkmark} & \multicolumn{1}{c|}{\checkmark} & \multicolumn{1}{c|}{} & \multicolumn{1}{c|}{} & \multicolumn{1}{c|}{} & \\ \cdashline{1-7} \cline{8-13}
Chen et al. \cite{chen_context-aware_2020} & 2020 & data sparsity - joint modeling of location and time prediction & DeepJMT for joint mobility and time prediction & Neural Networks & Foursquare & acc@ {[}5,10,20{]} & \multicolumn{1}{c|}{\checkmark} & \multicolumn{1}{c|}{\checkmark} & \multicolumn{1}{c|}{\checkmark} & \multicolumn{1}{c|}{} & \multicolumn{1}{c|}{} & \\ \cdashline{1-7} \cline{8-13}
Yang et al. \cite{yang_efficient_2020} & 2020 & data sparsity - low coverage & DestPD & Markov-based & Taxis GPS & Coverage - MAE & \multicolumn{1}{c|}{\checkmark} & \multicolumn{1}{c|}{\checkmark} & \multicolumn{1}{c|}{} & \multicolumn{1}{c|}{} & \multicolumn{1}{c|}{} & \\ \cdashline{1-7} \cline{8-13}
Comito \cite{comito_next_2020} & 2020 & integrating individual and collective mobility into prediction task - incorporate temporal context into prediction & NexT & Pattern-based & Foursquare - Twitter & Coverage - acc@ {[}1,10,20{]} - Overall performance & \multicolumn{1}{c|}{\checkmark} & \multicolumn{1}{c|}{\checkmark} & \multicolumn{1}{c|}{} & \multicolumn{1}{c|}{} & \multicolumn{1}{c|}{} & user and venue related context from frequent patterns of the geotagged tweets (e.g. identify night and weekend locations) \\ \cdashline{1-7} \cline{8-13}
Yang et al. \cite{yang_location_2020} & 2020 & data sparsity & Flashback & Neural Networks & Foursquare - Gowalla & acc@ {[}1,5,10{]} - Mean Reciprocal Rank (MRR) & \multicolumn{1}{c|}{\checkmark} & \multicolumn{1}{c|}{\checkmark} & \multicolumn{1}{c|}{} & \multicolumn{1}{c|}{} & \multicolumn{1}{c|}{} & spatial and temporal distance between each check-in \\ \cdashline{1-7} \cline{8-13}
Mo et al. \cite{Mo2021} & 2021 & extraction of latent activity patterns to infer travel purpose – predict stay time simultaneously & Input-output hidden Markov model (IOHMM) & Markov-based & Public transit smart card & Accuracy & \multicolumn{1}{c|}{\checkmark} & \multicolumn{1}{c|}{\checkmark} & \multicolumn{1}{c|}{} & \multicolumn{1}{c|}{\checkmark} & \multicolumn{1}{c|}{} & \\ \cdashline{1-7} \cline{8-13}
Sun et al. \cite{SunHeli2022} & 2022 & incorporate semantic features into prediction - cold start - data sparsity & ST-LSTM & Neural Networks & Foursquare & acc@ [1,5,10,50] - Recall & \multicolumn{1}{c|}{\checkmark} & \multicolumn{1}{c|}{\checkmark} & \multicolumn{1}{c|}{} & \multicolumn{1}{c|}{\checkmark} & \multicolumn{1}{c|}{} & \\ \cdashline{1-7} \cline{8-13}
Long et..al. \cite{Long2022} & 2022 & feature selection and extraction - data sparsity & Regularity and Preference-based Deep neural network (DeepRP) & Neural Networks & Vehicles GPS data & acc@ [1,5,10] - F1-Score - MRR & \multicolumn{1}{c|}{\checkmark} & \multicolumn{1}{c|}{\checkmark} & \multicolumn{1}{c|}{} & \multicolumn{1}{c|}{} & \multicolumn{1}{c|}{} & visit frequency\\ \cdashline{1-7} \cline{8-13}
Chen et al. \cite{Chen2022} & 2022 & Incorporate travel time context into prediction & Travel Time Difference Model (TTDM) & Graph Embedding / Probabilistic & Vehicles GPS data - Taxi trajectory data & acc@ [1,2.3.4.5] & \multicolumn{1}{c|}{\checkmark} & \multicolumn{1}{c|}{\checkmark} & \multicolumn{1}{c|}{} & \multicolumn{1}{c|}{} & \multicolumn{1}{c|}{} & \\ \cdashline{1-7} \cline{8-13}
\end{tabular}
\end{center}
\begin{enumerate}
   \item[*] No name is assigned to the proposed model
   \item[**] No model is proposed
\end{enumerate}
\end{table*}

\end{landscape}

% ========================== IV. Prediction Evaluation =========% 
% ==============================================================% 
\section{Data Sources and Evaluation Measures}
\label{datasets_and_evaluation_metrics}
Developing effective models for predicting the subsequent location of individuals or devices necessitates the use of appropriate datasets containing valuable information for both model training and evaluation. This section will provide an overview of the datasets frequently employed by various research studies, encompassing sources like GPS trajectory records, location-based social network (LBSN) check-ins, and detailed call records. These datasets play a crucial role in training and assessing next location prediction models. Subsequently, the following subsection will delve into the evaluation metrics commonly employed to gauge the performance and accuracy of such prediction methods.

\subsection{Datasets}
\label{Datasets}
Over the past few decades, advancements in technology and the Internet of Things (IoT) have facilitated the accumulation of substantial mobility data. This data is obtained either passively through embedded devices like GPS in vehicles or actively shared by users on various platforms, such as location-based social network (LBSN) check-ins and geotagged tweets \cite{Wu2018}. This paradigm shift has enabled researchers to tackle the challenge of predicting next locations. However, the public unavailability of some of these datasets makes it difficult to reproduce results. To address this, Wu et al. \cite{Wu2018, Zheng2015} have compiled a comprehensive set of human mobility datasets, and Luca et al. \cite{Luca2021} have curated a valuable repository (note that not all datasets are suitable for individual-level next location prediction due to some being group-level datasets like crowd flow data). In this section, we will categorize and summarize datasets explicitly employed by researchers in the field of next location prediction.

\subsubsection{Global Positioning System (GPS) Traces}
A typical GPS trace consists of tuples \((u, t, lat, long)\), where \(u\) represents the user's ID, \(t\) is the timestamp, and \(lat\) and \(long\) denote the latitude and longitude of a specific location, respectively. This type of data often requires preprocessing to mitigate noise and handle missing information. As GPS data is considered raw, preprocessing is essential to extract contextual details such as semantic labels of locations and temporal context (e.g., day of the week, hour of the day). This conversion from raw GPS data to a format compatible with context-aware models facilitates subsequent analysis. GPS traces are frequently captured at predefined intervals by embedded positioning devices, making this data collection passive in nature. Noteworthy GPS traces datasets commonly employed for next location prediction tasks include:

\emph{GeoLife dataset:} The GeoLife dataset was assembled as part of the Microsoft Research Asia GeoLife project \cite{noauthor_geolife_nodate}. Over a span of three years (from April 2007 to August 2012), 182 users contributed to this dataset. It encompasses 17,621 trajectories, covering a cumulative distance of approximately 1.2 million kilometers and a total duration exceeding 48,000 hours. Diverse GPS loggers and GPS-equipped phones were employed to capture these trajectories at varying sampling rates. Notably, 91\% of the trajectories were recorded using dense sampling strategies, such as intervals of 1 to 5 seconds or distances of 5 to 10 meters between consecutive points. This dataset comprehensively documents a wide spectrum of users' outdoor movements, encompassing daily routines like commuting to work or returning home, as well as recreational and sporting activities like shopping, sightseeing, dining, hiking, and cycling.

\emph{T-drive dataset:} The T-drive dataset originates from the T-drive project \cite{yuan_t-drive_2010}, offering a real-world and extensive collection of taxi trajectory data. This dataset encompasses a remarkable 580,000 taxi trajectories recorded within the city of Beijing. The trajectories collectively span a distance of 5 million kilometers and encompass a staggering 20 million GPS data points.

\emph{GeoPKDD dataset:} The GeoPKDD dataset is a compilation of trajectory data derived from cars fitted with GPS receivers, operating within the city of Milan over a span of one week. This dataset was curated as part of the GeoPKDD project \cite{giannotti_geopkdd_2009}, and it presents a comprehensive overview of vehicular movement patterns and trajectories within the urban context.

\subsubsection{(Location-Based) Social Networks (LBSN) Data}
Social media posts shared by users can either include or omit geographical information about their locations. Platforms like Facebook and Twitter, for instance, may contain posts with geographic details such as point of interest (POI) names or latitude/longitude coordinates. Conversely, platforms like Foursquare and Gowalla necessitate users to explicitly specify their locations in each shared post, leading to what is termed as geotagged social media posts. A geotagged post on social media refers to any content, be it text, photos, or videos, shared by a user. This content is associated with geographical information indicating the user's location. Depending on the platform, this location information might manifest as venue identifiers, categories, or geographical coordinates, along with a timestamp denoting the time of the post. Such a specialized form of social media, wherein all posts are required to be geotagged, is commonly referred to as Location-Based Social Networks (LBSNs). The general term used for user-shared content on LBSNs is "check-ins."

While data gathered from these LBSNs significantly contributes to the task of predicting the next location, other social networks also possess substantial amounts of geotagged content. The value of this data is further augmented by the spatiotemporal context present in the posts, which can include text, photos, and videos. Another notable aspect of social network data is the presence of friendship links between users. As previously mentioned, researchers have extensively explored how these social connections and relationships can impact human mobility and its predictability \cite{wang_regularity_2015, cho_friendship_2011}. However, this type of data is not without limitations. Given that social media posts are typically shared voluntarily by users, they fall under the category of actively recorded data. Users' locations are only recorded when they actively post something or check in at a location, which gives rise to the issue of data sparsity. To facilitate research, most social networks offer APIs for data retrieval, albeit often subject to restrictions such as a limit on the number of downloadable posts or queries per day. Commonly employed social network datasets for next location prediction tasks encompass platforms like Foursquare \cite{yang_revisiting_2019}, Twitter, Gowalla \cite{cho_friendship_2011}, Brightkite \cite{cho_friendship_2011}, Jiepang (a Chinese

\subsubsection{Call Detail Records (CDR)}
A call detail record (CDR) refers to a data record generated by a telephone exchange or similar telecommunications equipment. It serves to document the specifics of a communication event, encompassing calls and other telecommunication transactions such as text messages and data usage, that traverse through the said facility or device. These records play a pivotal role in various aspects, including billing and operational management. Telecom service providers maintain this data for diverse purposes.

In the context of CDRs, each user's connection to a Radio Base Station (RBS), responsible for covering a designated geographical area, assumes significance. Consequently, each CDR encapsulates pertinent details such as the identities of the involved users in the communication, the corresponding RBS assignment during the transaction, the timestamp of the communication event, and supplementary information. However, a notable drawback associated with this data type is the inherent data sparsity. User location information is only logged when users engage in telecommunication interactions, contributing to a scarcity of location data.

Another limitation pertains to the manner in which user locations are recorded within CDRs. As highlighted earlier, these records often log the ID or location of the RBS that facilitated the user's network connection, rather than the precise user location. This practice compromises the accuracy of the provided location information. Given the sensitive nature of the personal information contained within CDRs, they are generally not accessible to the public. In certain cases, access may be granted for research purposes, contingent upon approval. Nonetheless, this restricted availability hampers the reproducibility of results derived from using CDR data for training and evaluation purposes.

\subsection{Evaluation Metrics}
The next location prediction problem is approached by the analysed studies through two primary methods. The first method involves predicting the precise geographical coordinates of a user's next location. Here, the problem is framed as a regression task, with the target variables being the future longitude and latitude coordinates. Evaluating the performance of models in this scenario entails employing distance metrics. 

Conversely, the second method treats the problem as a classification task, aiming to predict the subsequent place, venue, or point of interest (POI) that the subject will visit. This approach is inherently more context-aware compared to the former, as it incorporates semantic attributes of the locations, venue categories, and the user's activities during their time at those locations.

Another subset of methods adopts a different approach by dividing the spatial map into grids of uniform size. The objective is to predict the specific grid cell where a user or device will be situated next. While this method falls under the classification-based approach, it shares similarities with regression-based techniques from the context-awareness perspective. 

\subsubsection{Distance Metrics}
For approaches treating the next location prediction problem as a regression task, the objective is to forecast the precise future location of the subject. The evaluation of these models is centered around the discrepancy between the predicted geographical location and the actual location \cite{cho_friendship_2011,yang_efficient_2020}. 

Distance metrics play a crucial role in this evaluation. Typically, the distance between two points can be quantified using either the Euclidean distance or the haversine distance, which calculates the angular separation between two points on the surface of a sphere. The haversine distance between two points \(p_1(\phi_1, \lambda_1)\) and \(p_2(\lambda_2, \lambda_2)\) is computed using Equation \ref{haversine}.

\begin{equation}
\label{haversine}
    distance_h = 2R\:arcsin(\sqrt{sin^2(\frac{\phi_2-\phi_1}{2})+cos\phi_1.cos\phi_2.sin^2(\frac{\lambda_2-\lambda_2}{2})})
\end{equation}

Here, \(\phi\) and \(\lambda\) represent the latitude and longitude of a point, respectively. After calculating the distance error for each prediction query, performance metrics like mean average error (MAE), mean squared error (MSE), or root mean squared error (RMSE) are employed to indicate the model's average error (Equations \ref{mae}, \ref{mse}, \ref{rmse}). However, it's important to note that comparing this metric across different users might introduce bias, as users with larger gyration radius tend to have higher average errors. To address this, the MAE can be normalized by dividing the average error of each user by their corresponding average gyration radius.

\begin{equation}
\label{mae}
    MAE = \frac{1}{|Q|}\sum_{i=1}^{Q}\:Distance(True,Prediction)
\end{equation}
\begin{equation}
\label{mse}
    MSE = \frac{1}{|Q|}\sum_{i=1}^{Q}\:Distance^2(True,Prediction)
\end{equation}
\begin{equation}
\label{rmse}
    RMSE = \sqrt{\frac{\sum_{i=1}^{Q}\:Distance^2(True,Prediction)}{|Q|}}
\end{equation}
\subsubsection{Classification and Ranking Metrics}
For approaches treating the next location prediction as a classification problem, the focus is on predicting the discrete venue or place the subject will visit next, or ranking the most probable venues. Several metrics are commonly used to evaluate these methods, each shedding light on different aspects of model performance \cite{Long2022,liu_predicting_2016}.

\noindent{\bf Precision/Accuracy:} Precision, also referred to as accuracy in some studies, reflects the model's precision. It is calculated as the ratio of correct predictions to all predictions made (Equation \ref{accuracy}) \cite{Chen2022,Mo2021}.

\begin{equation}
\label{accuracy}
    accuracy/precision = \frac{correct\:predictions}{predictions\:done} = \frac{p^+}{p^{+}+p^{-}}
\end{equation}

Here, \(p^+\) and \(p^-\) represent the number of correct and incorrect predictions. A variant of accuracy, known as acc@K, is used. In this case, a prediction is considered correct if the actual next location (ground truth) appears within the top-X predicted places.

\noindent{\bf Coverage and Recall:} Coverage, or prediction rate, is employed by both regression- and classification-based models to showcase the model's prediction ability. It's the proportion of prediction queries for which the model delivered a result (true or false) out of the total requested queries (\(|R|\)) (Equation \ref{coverage})\cite{monreale_wherenext_2009,xue_destination_2013,yang_efficient_2020}.

\begin{equation}
\label{coverage}
    coverage = \frac{predictions\:done}{predictions\:requested} = \frac{p^{+}+p^{-}}{|R|}
\end{equation}

Recall offers a comprehensive view of precision and coverage, calculated by dividing the number of correct predictions by the total number of predictions requested (Equation \ref{recall}).

\begin{equation}
\label{recall}
    recall = \frac{correct\:predictions}{predictions\:requested} = \frac{p^+}{|R|}
\end{equation}

Similar to accuracy, recall can be applied as recall@K to alleviate result stringency.

\noindent{\bf F-measure:} The F-measure, also known as F-score or overall performance, combines precision and recall. It's the harmonic mean of these two metrics (Equation \ref{fscore}) \cite{cho_friendship_2011,ying_mining_2013,liu_predicting_2016,Long2022}. The F-measure assesses a classifier's ability to predict accurately (correctly classifying instances) and robustly (not missing a significant number of instances). A high-precision, low-recall classifier is highly accurate but misses hard-to-classify instances. Conversely, a high-recall, low-precision classifier provides a prediction for every query but has mostly incorrect predictions. Some studies calculate F-measure@K using acc@K and recall@K (Equation \ref{fscore2}).

\begin{equation}
\label{fscore}
    F-measure = 2\:.\:\frac{precision\:.\:recall}{precision\:+\:recall}
\end{equation}

\begin{equation}
\label{fscore2}
    F-measure@K = 2\:.\:\frac{acc@K\:.\:recall@K}{acc@K\:+\:recall@K}
\end{equation}

\noindent{\bf Mean Reciprocal Rank (MRR):} The mean reciprocal rank measures processes that provide lists of possible responses to questions, ordered by how likely they are to be correct. When we look at the order of correct answers, the rank is given as a fraction: 1 for the first place, \(\frac{1}{2}\) for the second place, \(\frac{1}{3}\) for the third place, and so on \cite{Chen2022,yang_location_2020,kong_hst-lstm_2018}. The average of these fractions for a group of questions Q gives us the mean reciprocal rank. This measure is stricter than accuracy and acc@K because it considers both how well the correct answer is ranked and its actual rank.

\begin{equation}
\label{mrr}
    MRR = \frac{1}{|Q|}\sum_{i=1}^{Q}\frac{1}{rank_i}
\end{equation}

\noindent{\bf Average Percentile Rank (APR):} The average percentile rank is found by calculating the average of Percentile Rank (PR) scores for all user check-in predictions \cite{noulas_mining_2012,wang_regularity_2015}. It shows the average position of the correct answer in the list of ranked answers (Equation \ref{apr}). A PR score of 1 means the predicted place is ranked first, and the score decreases linearly to 0 as the correct place's rank goes down (N is the total number of possible places).

\begin{equation}
\label{apr}
    APR = \frac{1}{|Q|}\sum_{i=1}^{Q}\frac{N-rank_i+1}{N}
\end{equation}

% ========================== V. Discussion =====================% 
% ==============================================================% 
\section{Critical discussion}
\label{discussion}
\begin{figure}
    \centering
    \caption{Summary of contextual information and methods used for addressing the next locations problem}
        \begin{subfigure}[b]{0.4\textwidth}
            \centering
            \includegraphics[width=\textwidth]{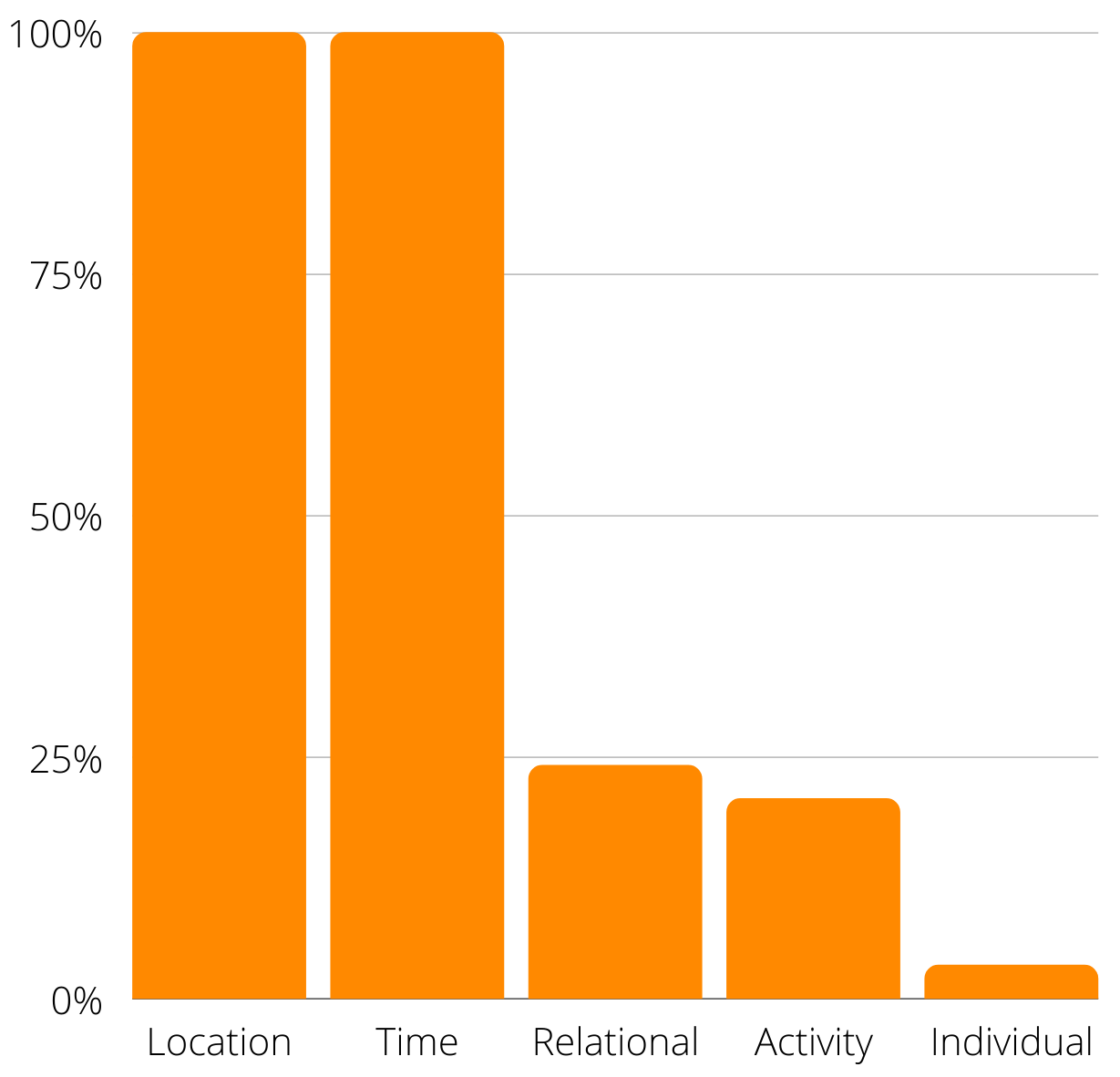}
            \caption{Proportion of papers that incorporated different context categories for next location prediction}
            \label{fig:contex_types_used_in_papers}
        \end{subfigure}
        \hfill
        \begin{subfigure}[b]{0.4\textwidth}
            \centering
            \includegraphics[width=\textwidth]{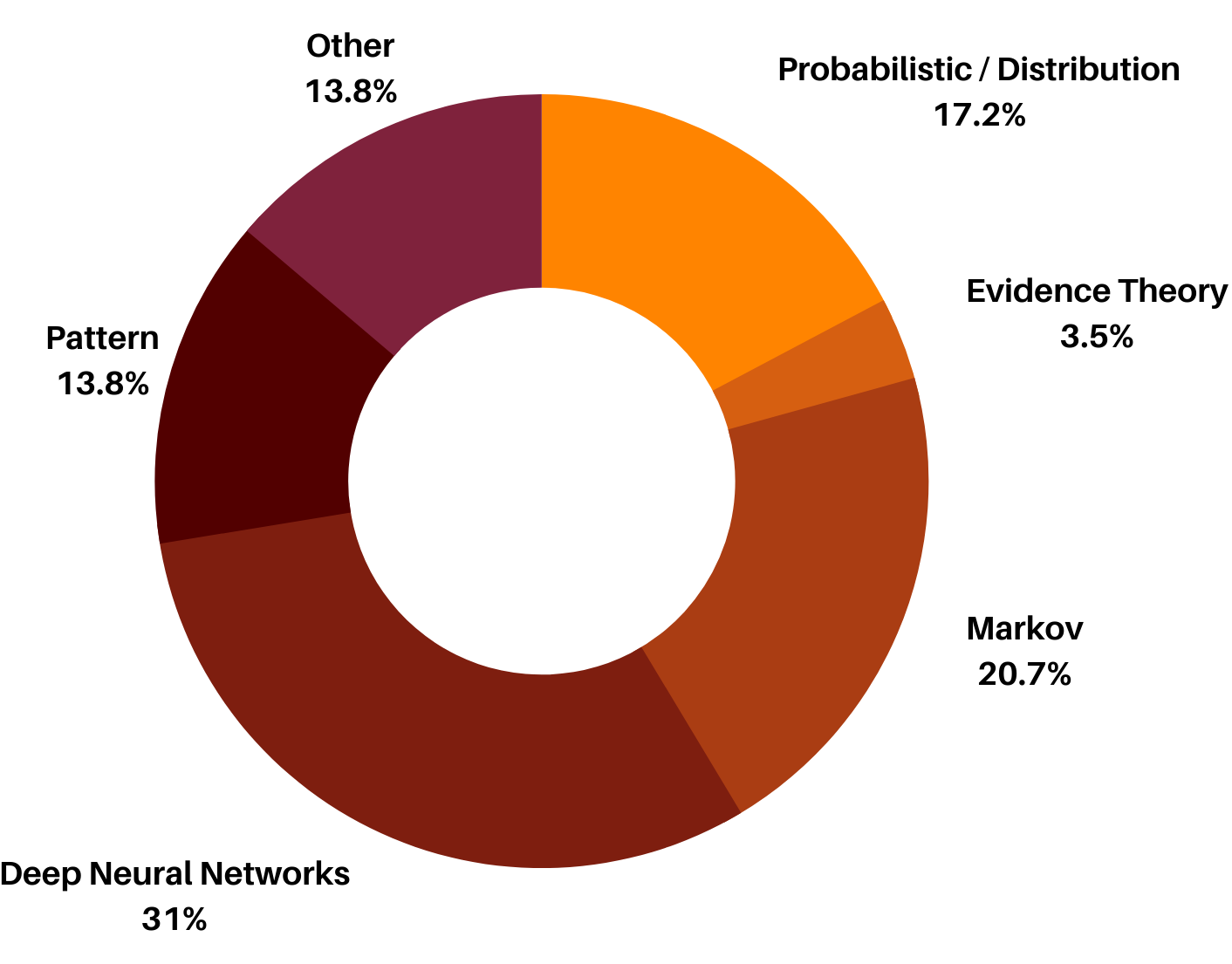}
            \caption{Proportion of different methods exploited for next location prediction}
            \label{fig:methods_in_papers}          
        \end{subfigure}
\end{figure}

In this section, we summarise and discuss the findings of this survey. Figure \ref{fig:contex_types_used_in_papers} demonstrates the proportion of papers that used each type of context. According to this chart, all the papers we analysed incorporate spatiotemporal context for predicting the next location. Less than 25\% of the studies used activity and relational context. Individual context (containing information about the entity's profile containing his schedule, tasks, habits, etc.) was used by only one study.

Nevertheless, these neglected types of contextual information need to be exploited to predict the next location. Involving activity context (venue category) contributes to predicting the stay time at a place as well as elevating the prediction accuracy \cite{Mo2021, SunHeli2022, yang_revisiting_2019}. The same as activity context, relational context (it's called {\emph social ties} by some studies) can improve the prediction accuracy \cite{cho_friendship_2011, gao_exploring_2012, noulas_mining_2012}. Nearly 25\% of the papers we analysed use social ties between different users and knowledge about a user's friends or social relatives and show that this information can help predict the user's future whereabouts. Social relationships can explain about 10\% to 30\% per cent of mobility behaviour \cite{cho_friendship_2011}. Another research \cite{gao_exploring_2012} exploring the effect of social-historical ties on LBSNs, shows that, on average, two strangers share roughly four check-ins, while two friends on an LBSN share nearly 12 check-ins which indicates the correlation of mobility behaviour between related entities.

Presumably, individual context is the most neglected type of context in human mobility prediction. As shown in Table \ref{tab:table1} and Figure \ref{fig:contex_types_used_in_papers}, \cite{samaan_mobility_2005} is the only study that takes individual context into account (user tasks, schedule, interests). Incorporating individual context raises data collection/acquisition and privacy challenges. Despite the spatiotemporal context, which can be collected easily by smartphones or GPS devices, individual context is difficult to gather (e.g. collecting a user's appointments, tasks, current health situation). Accessing individuals' personal information can lead to potential privacy leak threats and being used by malicious parties raising the privacy preservation issue.

In the case of predicting a vehicle's next location, the main entity is the vehicle itself, and all the contextual information about the other entities related to the vehicle is considered to be relational context. Since vehicles cannot make decisions, the most important entities to be considered are the drivers/passengers. So knowing about the driver/passengers' profile, preferences, intentions, and mobility patterns is vital for the prediction task. Besides, there are other entities related to a vehicle that their contextual information and situation should be taken into account when predicting the vehicle's next location (e.g. environmental conditions, restrictions in traffic control of a city).

Regarding prediction methods exploited for the next location prediction task, there are some pros and cons to each method category. Figure \ref{fig:methods_in_papers} demonstrates the exploitation ratio of different prediction methods by the papers we analysed in the previous section. Figure \ref{fig:contex_types_used_in_papers} demonstrates the share of each method category in the studies we analysed. Nearly 30 per cent of the articles exploited DL-based approaches to tackle the next location prediction task, followed by Markov-based, probabilistic/distribution-based and pattern-based methods by 20, 17 and 14 per cent, respectively. In the rest of this section, we will discuss the advantages and drawbacks of each of these methods.
Distribution and probabilistic based approaches rely on the temporal aspect of movements and predict the probability of each location to be visited next, based on the temporal occurrence of visits in the past \cite{cho_friendship_2011}. These techniques model mobility behaviour by considering location, time, and other variables as independent variables. Although probabilistic theories such as Bayes Theorem provide foundations to accumulate evidence from existing context and ignore it in case of missing evidence, the independency assumption between different variables limits their prediction ability.

Markov model-based methods using the Markov property (the future location of an entity depends only on its current or last several locations) cannot capture the long-term dependency of regular movements and only consider the short-term history in predicting the next location. These models do not utilise the temporal context fully and only use it to extract the sequence of transitions between different locations. Markov model-based approaches share these two disadvantages with pattern-based methods.

Traditional machine learning (ML) approaches demand substantial efforts in feature engineering, a complex and time-intensive procedure reliant on domain expertise. Furthermore, this process may inadvertently overlook valuable features. In contrast, deep learning (DL)-based methods have demonstrated encouraging performance across various domains, including spatiotemporal challenges. Leveraging their distinctive capacity for automatic feature selection and extraction, coupled with their adeptness at discerning intricate patterns and approximating complex functions, Deep learning (DL) methods can proficiently grasp the spatial, temporal, social, and geographic aspects of human mobility. This is achieved through the utilization of Convolutional Neural Networks (CNNs) for spatial context, Recurrent Neural Networks (RNNs) for temporal context, and attention mechanisms for capturing long-term dependencies.

However, a considerable amount of data is required to enable a DL-based model to extract appropriate features and successfully capture complex patterns. The lack of enough data for training the model brings down their performance extensively, which is their main drawback compared to their rivals. It also requires expert knowledge for designing the proper network architecture and tuning hyperparameters. Another drawback to these models is that they are less interpretable than the traditional ML approaches.

% ================ VI. Challenges and Future Work=============% 
% ============================================================% 
\section{Open challenges and future work}
\label{future_work}
This section will discuss the open challenges and possible future work in the next location prediction domain. 
\subsection{Data collection}
One of the challenges that future research on this subject faces is data collection and acquisition. As explained before, most methods are dependent on data for their training and evaluation phases. Researchers rely on existing datasets such as GPS traces, CDRs and LBSN check-in datasets; however, the contextual information provided by these datasets is limited, making it difficult to extract meaningful insights and semantics. Although data collected from LBSNs contains location, time, activity and relational context, this data source suffers from the lack of useful individual context and data sparsity problems. Unlike GPS traces and CDR data, which are collected regularly, LBSN posts are shared actively by the users (voluntarily). There may be workarounds to this challenge. One solution is trying to collect large-scale semantically enriched information \cite{laurila2012mobile}. Another solution is fusing contextual information from multiple data sources and accumulating existing evidence that helps achieve a better prediction. \cite{samaan_mobility_2005, wang_regularity_2015}.
\subsection{Randomness and prediction accuracy}
Although mobility behaviour follows some patterns and humans usually commute based on regular patterns, there is inherent randomness in it to a certain level. According to the findings of \cite{song2010limits} the maximum predictability of human mobility can reach up to 93\% if all the regularities and patterns are adequately captured, and the remaining 7\% follows a random, stochastic and unpredictable pattern. Cho et al. \cite{cho_friendship_2011} shows that periodic mobility behaviour of the users explains 50\% to 70\% of all human movement while social relations can explain 10\% to 30\% of it. As another exploration into human mobility, Cuttone et al. \cite{cuttone2018understanding} report that for each user, on average, only 5\% to 10\% of the places are visited more than once, and they visit 70\% of the venues only once and 20\% to 25\% of the destinations are new places. All these findings point to the importance of studying the {\emph exploring} attitude of human mobility behaviour alongside mining regularity patterns. One solution to tackle this challenge is predicting a user's intention to explore before attempting to find his/her future location based on his/her movement pattern. This issue can be addressed by combining the problem of movement pattern mining which tries to capture the regularities in movement behaviour, with the problem of the (next/sequential) POI recommendation, which focuses on finding the new venues that probably the user will be interested in visiting. 
\subsection{Privacy preservation}
The analysis of mobility behavior offers both opportunities and privacy risks to users. Existing methods addressing the next location prediction challenge require access to diverse contextual information linked to users to achieve accurate predictions. This information encompasses location, time, social relationships, activities, and personal data collected from GPS devices, smartphones, and social media platforms. Consequently, ensuring privacy preservation becomes of paramount importance for individuals utilizing these services \cite{xue_destination_2013, feng_pmf_2020}. Such information could empower malicious entities to deduce users' mobility patterns, discern their residential and workplace locations, and construct detailed models of their routine movements and social networks. These privacy threats extend beyond mere knowledge of an individual's whereabouts. With the digitization of vehicles gaining momentum, the susceptibility to cyberattacks has increased, elevating the significance of privacy preservation. Unauthorized access to users' personal information, particularly their location and daily routines, amplifies the potential for malicious actors and heightens the risk of successful cyberattacks.
\subsection{Cold start and data sparsity}
Various studies addressed the problem of cold start; however, some aspects of this problem remain unsolved. The cold start issue arises when there is no or few historical data for a user. In this case, the standard approach is to match the user's current trajectory's pattern (since it may be the only spatiotemporal information in hand) with similar other users may give us intuition about predicting the user's next destination. Individual contextual information can play a crucial role in tackling the cold start problem. This type of context refers to the information about the entity itself and is independent of time and location, so it changes with less frequency. Incorporating this context into the prediction task can help recognise the user's personality, which may lead to a better guess about their movement patterns. As a future direction to address this issue, rule-based methods can also be considered.

The data sparsity problem concerns the situation where the spatiotemporal data is not dense enough to illustrate the whole movement patterns of the users. The sparsity depends on the source of data used for training and inference. As discussed in section \ref{Datasets}, GPS trajectory sources are sufficiently dense, but noise filtering, extracting semantics and meaningful concepts from this type of data remains a challenging task. On the other hand, LBSN data are semantically rich; however, since they are recorded voluntarily by the users, this data source suffers from the data sparsity problem the most. CDR data sources also lack adequate denseness because they only contain records of telecommunication transactions. Also, the location data in these datasets are not accurate enough to enable semantics extraction since they log the locations of RBSs instead of the users. Overall, the GPS trajectory data seems to be the best data source for this task. However, semantic information should be extracted to obtain contextual information about the places and achieve context-aware predictions.
\subsection{Future work}
An unforeseen yet crucial application of mobility prediction is poised to emerge as over-the-air (OTA) updates for vehicles become increasingly prevalent in our daily lives. OTA updates for cars can be categorized into two primary types: infotainment and drive control. Infotainment updates aim to improve the in-car experience and are not vital for the vehicle's operation. In contrast, drive control OTA updates directly affect how a vehicle functions safely while driving. These updates involve enhancements or fixes to essential systems like powertrain, chassis, advanced driver assistance systems (ADAS), and security features, which are crucial for the vehicle's proper functioning.

OTA updates are delivered automatically in the background while the vehicle is in use and connected. When the driver stops and turns off the ignition, a notification appears on the infotainment system or mobile app, informing the user about the update and seeking their permission for installation. To ensure safety and mitigate potential issues, the vehicle must be parked and the ignition switched off for the installation to commence. This dynamic introduces the challenge of finding a balance between driver/passenger convenience and security/safety. In the context of OTA updates, solely predicting the next location or destination of the vehicle falls short of meeting the requirements, given the need for safety, stability, and other considerations. Instead, predicting the next "optimal" location or destination assumes paramount importance and urgency, characterized by the following aspects:

\begin{itemize}[noitemsep,nolistsep]
\item {\bf Adequate stay time:} There will be adequate time for the installation to take place.
\item {\bf Reliability:} The place where the OTA updates are applied should be a safe and reliable place to prevent or reduce the safety risks in case of failure or malfunctioning. 
\item {\bf Connectivity:} The place where the updates are installed needs to have internet connectivity to make it possible for the vehicle to re-download the update packages in case of corruption or incompatibility. 
\end{itemize}
    
Predicting a vehicle's next location also holds significance for the automotive industry in aiding threat analysis and risk assessment regarding cybersecurity attacks. More specifically, it assists in estimating the window of opportunity for potential attacks. The ISO/SAE 21434:2021 standard \cite{iso21434}, which pertains to cybersecurity engineering for road vehicles, categorises the window of opportunity as "Unlimited," "Easy," "Moderate," and "Difficult" based on factors such as physical access and available time for an attack on the vehicle. Anticipating a vehicle's future location(s), closely linked to the locations of its driver and passengers, aids in identifying the window of opportunity category and evaluating associated risks. For instance, predicting that a car will be inside a car wash or repair garage implies a significant exposure time for the vehicle. Likewise, if the prediction indicates that the vehicle will be parked near a shopping mall, it signals exposure to diverse connectivity surfaces such as cellular, WiFi, and Bluetooth.

% ================ VII. Conclusion ===========================% 
% ============================================================%
\section{Conclusion}
\label{conclusion}
This paper performs a comprehensive overview of recent advances in addressing the next useful location prediction problem by examining the context-awareness aspect. Twenty-nine studies in this field are analysed and categorised based on their methods, the contextual information incorporated, the challenges addressed, and the datasets used. These method categories consist of evidence theory-based, neural networks-based, pattern mining-based, distribution, probabilistic-based and hybrid methods. An operational definition of contextual information comprising individual, location, time, activity and relational is considered to examine these studies from the context-awareness perspective. We discuss the strengths and weaknesses of the existing methods focusing on their ability to employ different context categories. Finally, we listed some of the open problems and pointed out the future research directions for this field. We also introduce two automotive cybersecurity-related use cases for the next location prediction task.

% ================ VII. References ===========================% 
% ============================================================%
\bibliographystyle{unsrt}

\end{document}